\documentclass[12pt]{article}

\usepackage{amsmath,amssymb,bm}
\usepackage{graphicx}
\usepackage{caption}
\usepackage{color, cancel, ulem, soul}
\usepackage{bm}
\usepackage{float}

\usepackage{amsmath,amssymb,graphicx}
\usepackage[document]{ragged2e}

\usepackage{bm}
\newcommand{\bx}{\bm x}
\newcommand{\bw}{\bm w}
\newcommand{\bzero}{\bm 0}
\newcommand{\bI}{\bm I}
\newcommand{\bmu}{\bm\mu}

\usepackage{times}
\usepackage{bm}
\usepackage[linesnumbered,lined,boxed,commentsnumbered,ruled,vlined]{algorithm2e}

\title{Nonparametric Density
Estimation for High-Dimensional Data - {Algorithms and Applications}}

\author{Zhipeng Wang\thanks{Department of Statistics, Rice University, Houston, TX, email: zw12@rice.edu} \ and David W.~Scott\thanks{Department of Statistics, Rice University, Houston, TX  77251-1892, email: scottdw@rice.edu.
}}

\date{}

\begin{document}
\maketitle

\begin{center}
\subsubsection*{\small Article Type:}
{\color{black}Advanced Review}

\hfill \break
\thanks

\subsubsection*{Abstract}
\begin{justify}

 Density Estimation is one of the central areas of statistics whose purpose is to estimate the probability density function underlying the observed data. It serves as a building block for many tasks in statistical inference, visualization, and machine learning. Density Estimation is widely adopted in the domain of unsupervised learning especially for the application of clustering.  As big data become pervasive in almost every area of data sciences, analyzing high-dimensional data that have many features and variables appears to be a major focus in both academia and industry. High-dimensional data pose challenges not only from the theoretical aspects of statistical inference, but also from the algorithmic/computational considerations of machine learning and data analytics.  This paper reviews { a collection of selected nonparametric density estimation algorithms for high-dimensional data, some of them are recently published and provide interesting mathematical insights.} {The important application domain of nonparametric density estimation, such as { modal clustering}, are also included in this paper}. { Several research directions related to density estimation and high-dimensional data analysis are suggested by the authors. }

\end{justify}
\end{center}

\noindent
{\bf Keywords: } 
Density Estimation, High-dimensional data, Clustering, Neural Networks, Data Partitioning

\clearpage


\renewcommand{\baselinestretch}{1.5}
\normalsize

\clearpage

\section*{\sffamily \Large GRAPHICAL TABLE OF CONTENTS} 
Include an attractive full color image for the online Table of Contents. It may be a figure or panel from the article, or may be specifically designed as a visual summary. You will need to upload this as a separate file during submission.

\begin{figure}[H]
	\centering
	\setlength{\fboxsep}{15pt}
	\setlength\fboxrule{4pt}
	\fbox{\includegraphics*[width=0.95\textwidth]{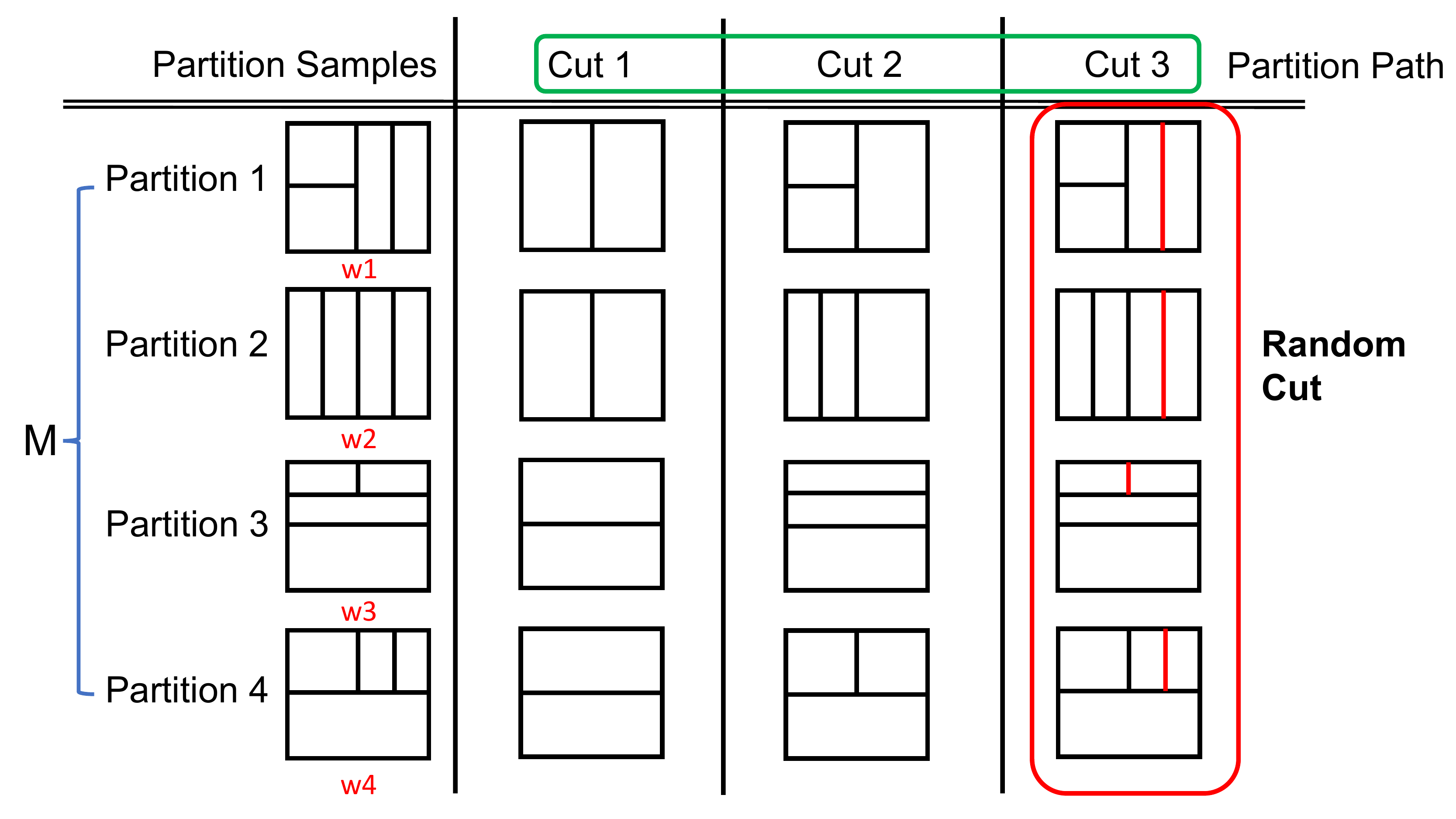}}
	\label{fig:0}
\end{figure}



\clearpage

\section*{\sffamily \Large INTRODUCTION} 
\begin{justify}
Density Estimation is a widely adopted tool for many tasks in statistical inference, machine learning, visualization, and exploratory data analysis.  The aim of density estimation is to approximate the probability density function underlying the data, which are assumed to be $i.i.d$.  Existing density estimation algorithms can be categorized into either parametric, semi-parametric, or nonparametric approaches. Parametric density estimation algorithms are model-based, usually come with strong assumptions on the distribution of the underlying data. One of the most widely-adopted parametric or semi-parametric density estimation algorithms is the Gaussian Mixture Model (GMM), which was first considered by Karl Pearson (1894) \cite{Pearson:1894} followed by many recent works including Aitkin and Wilson (1980) \cite{Aitkin:1980}, Hathaway (1985) \cite{Hathaway:1985}, McLachlan and Krishnan (2008) \cite{McLachlan:2008}, and Wang and Wang (2015) \cite{Wang:2015}.  The basic idea is that given $i.i.d.$ data $\bx_i \in \Re^d$ , the probability distribution of $\bx$ can be modeled as a simple linear superposition of Gaussian Components: \\
\begin{equation}
\hat{f}_{K}(\textbf{x}) = \sum_{k=1}^{K} w_{k}\, \phi (\textbf{x}|
\bmu_k, \Sigma_k)\,,
\label{eq:GMM}
\end{equation}
where the nonnegative weights $w_k$ sum to 1. Choosing the number of Gaussian Component $K$ is usually a tricky task. Theoretically the Gaussian Mixture model can estimate any density function if $K$ is large enough. In practice, however, increasing $K$ would lead to large numbers of parameters to be estimated by the maximum likelihood algorithm. Since $\bx \in \Re^d$, the most general model contains $K-1$ parameters in the weight vector, $K\times d$ parameters in the mean vectors, and $K\times d(d+1)/2$ parameters in the covariance matrices. This will result in computational challenges and more importantly, will require a much bigger dataset. In the application of clustering, each component in the Gaussian Mixture model naturally corresponds to one cluster, and one expects in the ideal case that the $K$-component Gaussian Mixture Models would illustrate $K$ modes in the density function.  {Unfortunately, this is not always the case.} This problem was pointed out in previous research such as Good and Gaskin (1980) \cite{Good:1980} and
       Roeder (1990) \cite{Roeder:1990}.  Another natural challenge is the choice of initial values of the parameters for a maximum likelihood algorithm. If one assumes $\Sigma_k = \Sigma$ for all $k$, then the number of parameters will be significantly reduced at the cost of generality and possibly accuracy, even when $K$ is dramatically increased. One problem is that if we make the assumption of a fully general covariance matrix and if the determinant of any one of the $\Sigma_k$ approaches 0, then the maximum likelihood
criterion will approach $+ \infty$. However, theoretical results and practical experience
show that there are many local maxima in the likelihood function that provide useful estimates (Hathaway, 1985 \cite{Hathaway:1985}). Thus trying a number of different initializations is highly recommended. (Anecdotally, the initialization problem is considered to be NP-hard in the machine learning literature.)  When applying Gaussian Mixture Modeling, there is considerable interest in the relative size of the components. The estimated weights ${\hat{w}_k}$ provide a natural choice for this purpose. However, the Gaussian density does not provide an orthonormal basis for density functions. In the $L_2$ function space, the mixture model is dense but the true number of components $K$ might be quite large (or even infinite). However, there are many solutions where the individual components $\{w_k,\mu_k,\Sigma_k\}$ are quite different, but the overall
sums in Equation \eqref{eq:GMM} are visually identical \cite{Cleveland:1993, Cleveland:1994}.  This is the result of the basis not being orthogonal, so that there is high correlation among the estimated parameters in the GMM. \\

\hspace*{8mm} We also note that there is a vast amount of literature on probabilistic graphical models (PGMs). 
PGMs are a parametric family of statistical models for approximating multivariate joint probability distributions using graphs to express the conditional dependence structure between random variables. They are widely used in probability theory and
statistics---especially in Bayesian statistics and machine learning.
Useful references are { Jordan (2004) \cite{MJordan:2004},
Koller (2009) \cite{Koller:2009},
and Wainwright (2010) \cite{Wainwright:2008}.}
Graphical models constitute
an active area of research in machine learning and statistics, which we will not cover in this article. Interested readers should refer to the references mentioned above or
delve into some relevant extensions and applications; see {
Wand (2016) \cite{Wand:2017},
Minka (2005) \cite{Minka:2005}, and
Yedidia et al.~(2003) \cite{Yedidia:2003}}. \\

\hspace*{8mm} The intrinsic problems arising from parametric PDE approaches promote the development of nonparametric density estimation. {In this article we will cover some interesting nonparametric density estimation algorithms. Especially we introduce the algorithms that are potentially suited for high-dimensional data. Here we define the "high-dimensional data" as the data with $3 < d <\leq 50$, where $d$ is the number of dimensions. We understand that it is rather a subjective concept and might have different range given different problems. Our definition of high-dimensional data was motivated by the work of Wong \cite{LuoWong:2013, Wong:2014} which mentioned that the ideal density estimator should be able to reliably estimate density functions for high-dimensional data with dimensions from 4 to 50. }   Nonparametric methods provide powerful and flexible means to estimate density functions, and thus have become a very active research topic in the field. Existing nonparametric density estimation algorithms include histograms \cite{Scott:1979}, frequency polygons \cite{Scott:1985}, Kernel Density Estimation \cite{Scott:1977, Silverman:1986}, Splines \cite{Stone:1994, Eilers:1996}, and neural network-based density estimation \cite{Magdon:1998, Larochelle:2011, Uria:2013, Uria:2014, Uria:2016, Papamakarios: 2017}. This field is rapidly developing and {new techniques are being created} to address the pressing need of big data analytics.  They serve as a foundation for many applications such as clustering, which we will also discuss in this paper. \\

\hspace*{8mm} Clustering is one of the most important and challenging problems in machine learning.
It may be formulated as an unsupervised learning algorithm in which the class labels are unknown, not even the number of classes $K$. It has wide applications in data compression \cite{RCilibrasi:2005, YMarchett:2017}, anomaly detection \cite{Aliu:2012, TIwata:2016}, recommendation systems and Internet of Things (IoTs) \cite{Pham:2011, BreeseUAI:1998, Su:2009, TLopez:2011}, etc. Density estimation serves as a foundation for clustering, as one can find modes in the estimated density function,
        and then associate each mode with a cluster. 
The modal value itself is taken as the prototypical member of the cluster.
The resulting ``mode clustering" or ``modal clustering'' has been extensively studied;
see Carmichael et al.~(1968) \cite{Carmichael:1968}
and Hartigan (1975) \cite{Hartigan:1975} for seminal works, as well as
(Azzalini and Torelli (2007) \cite{Azzalini:2007},
Cheng (1995) \cite{Cheng:1995},
Chazal et al.~(2013) \cite{Chazal:2013},
Comaniciu and Meer (2002) \cite{Comaniciu:2002},
Fukunaga and Hostetler (1975) \cite{Fukunaga:1975},
Li et al.(2007) \cite{LiJia:2007}, 
Chac\'on and Duong (2013) \cite{Chacon:2013},
Arias-Castro et al.(2013) \cite{Arias:2013},
Minnotte and Scott (1993) \cite{Minnotte:1993},
Minnotte, Marchette, and Wegman (1998) \cite{Minnotte:1998}, and
Chac\'on (2016) \cite{Chacon:2016}.
The method might generate a conservative result in the sense that pairs of adjacent clusters might manifest as a single mode (or a single bump) in the kernel estimate.
But clustering is an exploratory activity, so such limitations should be tolerated.
Adding more informative variables might {help further separation of the clusters in the high dimensional feature space.} \\

Recent work by Chen (2016) \cite{Chen:2016}
provides several enhancements over the existing mode clustering formulations, including a soft variant of cluster assignments, a measure of connectivity between clusters, a method to denoise small clusters and a way to visualize clusters.  
A comprehensive survey of modal clustering has recently been provided by
Menardi (2016) \cite{Menardi:2016}, which should be read in parallel with
material below.
\\	

\hspace*{8mm} In addition to the approaches introduced above, there are many clustering algorithms that do not rely on a parametric or nonparametric probability density estimation of the data.
The most commonly used is the hierarchical clustering algorithm, which is implemented based on an iterative distance-based approach;  see Johnson (1967) \cite{Johnson:1967} and a recent overview in Izenman (2008)
\cite{Izenman:2008}.
The results of the algorithm are usually displayed as a binary tree. The most widely used nonhierarchical clustering algorithm is $k$-means
(MacQueen (1967) \cite{MacQueen:1967}) that iteratively updates the centroids of points currently assigned to the $k$ groups, then reallocates points to the closest centroid, and stops when no further updates occur. Recent work done by Chi and Lange (2015) \cite{ChiLange:2015} and Chi et al.~(2017) \cite{ChiAllenBaraniuk:2017} further extended the $k$-means and hierarchical clustering algorithms by proposing splitting algorithms for convex clustering problems. The basic idea is to formulate clustering tasks as a convex optimization problem, in which there is a unique global minimizer for the objective function and the cluster centroids are shrunk toward each other. Then a variety of splitting algorithms such as alternating direction method of multipliers (ADMM) and alternating minimization algorithm (AMA) can be adopted to solve the optimization problem;  see
       Chi et al.~(2015) \cite{ChiLange:2015}. \\




\hspace*{8mm} The remainder of this article will be divided as follow: we will first review some of the important algorithms in nonparametric density estimation, including neural networks-based density estimation algorithms as well as density estimation algorithms based on adaptive data partitioning and Projection Pursuit. Then we will switch our focus to mode clustering methods using nonparametric density estimation. Finally, we will provide critical comments on the limitations of density-based algorithms and suggest future research directions. 
 \\

{\section*{NONPARAMETRIC DENSITY ESTIMATION FOR \\ HIGH-DIMENSIONAL DATA} }

\hspace*{8mm} In the following sections, we will introduce relevant nonparametric algorithms for high-dimensional density estimation. Since neural networks gained popularity in recent years, we want to cover some relevant density estimation algorithms based on neural networks.  We will also introduce algorithms of Multivariate Density Estimation via adaptive sequential data partitioning, which were proposed by Luo and Wong (2013) \cite{LuoWong:2013} and Li and Wong (2016)\cite{LiWong:2016}. These density estimation algorithms provide both computationally efficient and statistically robust means for function estimation. {Projection Pursuit Density Estimation (PPDE), which was first introduced by Friedman and Tukey \cite{Friedman:1974} has evolved into an efficient method of density estimation for high-dimensional data}. We will also discuss the PPDE algorithm in this section. 


\subsection*{Density Estimation Based on Neural Networks}
One of the problems in kernel density estimation is that small changes of data and smoothing parameters can lead to large fluctuations in the estimated density. In order to make the estimation more robust to the slight changes of data, some regularization is usually needed. The regularizations are often reflected by choosing the smoothing parameters (kernel width or number of kernel functions $K(\cdot)$). However, the estimated density will be extremely sensitive to the choice of the smoothing parameter. A poor choice can lead to either oversmoothing or undersmoothing, either globally, locally, or both. \\

\hspace*{8mm} {The method of Neural Networks has recently gained tremendous popularity} in the machine learning community. It provides a powerful capability to estimation any function to any given precision while maintaining the flexibility to choose an error function to fit into the application. Neural network consists of many interconnected neurons, each neuron performs a nonlinear feature mapping $\sigma ({W}^T \bx + b)$, where $\bx \in \Re^d$ is the input data, $W$ is the weight vector for the neuron, and $\sigma$ is the nonlinear function (which is usually implemented as either sigmoid or ReLU function in practice \cite{Goodfellow-et-al-2016}). The underlying intuition is that the neural network can somehow learn the abstract representation of data by the exhaustive nonlinear mapping of the original features. Density estimation using neural networks once was used
very sporadically due to the limitation of computing resources. Magdon-Ismail and Atiya (1998) \cite{Magdon:1998}
proposed two methods of density estimation that can be implemented using multilayer neural networks. One is the stochastic learning of cumulative distribution function, which only works for univariate density estimation. Let $x_n \in \Re^1$,
where $n = 1,...,N$ and the underlying density is $g(x)$. Its cumulative distribution function is $G(x) = \int_{-\infty}^{x} g(x') \,dx'$. The density of the random variable $G(X)$ is uniform in $[0,1]$. Let the network output be denoted by
$H(x,w)$. The aim is to have $H(x,w) = G(x)$.
The basic idea behind the algorithm is to use the $N$ original data points as the input, 
and at each iteration cycle, new data points that are generated from a uniform distribution on the interval of $[0 ,1]$ as the network targets. The weights are then adjusted to map the new data points. Thus the neural network is trained to map the data to a uniform distribution. The algorithm is illustrated as follows: \\

\begin{enumerate}
	\item Let $x_1 \leq x_2 \leq ... \leq x_N$ be the data points. Set the iteration number of the training cycle $t = 1$. Initialize the weights of the neural network randomly to $w_1$. 
	
	\item Randomly generate N data points from a uniform distribution in $[0,1]$, and sort them in ascending order $u_1 \leq u_2 \leq ... \leq u_N$. Those points $u_n$ are the target output for the neural network with input $x_n$
	
	\item Update the network weights according to the backpropagation scheme: \\
	   $$w_{t+1} = w_t - \eta_t \frac{\partial J}{\partial w}\,,$$ \\
	   where $J$ is the objective function, and $\eta_t$ is the learning rate at each iteration.  The objective function $J$ includes the error term and the monotonicity penalty term: \\
	   $$J = \sum_{n=1}^N \big[{ H(x_n, w)} - u_n\big]^2 + \lambda \sum_{k=1}^{N_h} \Theta \big({ H(y_k, w) - H(y_k + \Delta, w)}\big) \big[{ H(y_k, w) - H(y_k + \Delta, w)}\big]^2\,.$$ \\
	   The first term is the standard L-2 error term, and second term is the monotonicity penalty term, $\lambda$ is a positive weighting constant, $\Delta$ is a small positive number, $\Theta (x)$ is the familiar unit step function, and the $y_k$ are any set of data points where the monotonicity is enforced. 
	   
	\item Move to the next iteration cycle $t = t + 1$ , and go to {step 3}. Repeat the process until the error is small enough. The resulting density estimate is the derivative of $H$.    
	   	     
\end{enumerate}

Another method they introduced is the smooth interpolation of the cumulative distribution (SIC),  which works for multivariate density estimations. The basic idea is that given the input data point $\bx \in \Re^d$, if the ground truth density function is $g(\bx)$, then the network target output is the corresponding cumulative distribution $\hat{G}(\bx)$. Let $\bx = (x^1, ..., x^d)^T$, $G(\bx)$ is given by: \\
\begin{equation}
G(\bx) = \int_{-\infty}^{x^1} ... \int_{-\infty}^{x^d} g(\bx)\, dx^1 \ldots x^d \,.
\end{equation}
Then we can approximate $G(\bx)$ by using the fraction of data points falling in the area of integration: 
\begin{equation}
\hat{G}(\bx) = \frac{1}{N} \sum_{n=1}^{N} \Theta (\bx - \bx_n)
\end{equation}
where $\Theta$ is defined as: \\
\begin{equation}
\Theta(\bx) = \left\{
\begin{aligned}
1 &   & \ \ x^i \geq 0\ \  (i = 1,2,...,d)\\
0 &   & \mbox{otherwise}\,.\qquad\qquad\\
\end{aligned}
\right.
\end{equation}
The $\hat{G} (\bx)$ is an estimate of $G(\bx)$
that is used for the target outputs of the neural network. The neural network model provides a smooth interpolation of the cumulative distribution function which is highly desirable. The density function is then obtained by differentiation of the network outputs with respect to its inputs. \\

For low-dimensional problems, we can do uniform sampling in (3) using a grid to empirically obtain examples for the target output of the network. For high-dimensional problems beyond two or three dimensions, the uniform sampling becomes computationally expensive. The alternative option is to use the input data points to form examples. To illustrate this, the target output for a input point $\bx_m$ would be: \\
\begin{equation}
\hat{G} (\bx_m) = \frac{1}{N-1} \sum_{n=1, n\neq m}^N \Theta (\bx_m - \bx_n)\,.
\end{equation}

Finally the monotonicity of the cumulative distribution function can be used as a hint to guide the training process. The network output $H(\bx, w)$ approximates the cumulative distribution function $G(\bx)$, then the density estimate can be derived as: \\
\begin{equation}
\hat{g}(\bx) = \frac{\partial^d H(\bx, w)}{\partial x^1...\partial x^d}\,.
\end{equation}


There are a variety of choices for the neural network architecture.  Feedforward neural network is commonly adopted (including both single and multiple hidden layers) \cite {Magdon:1998}; however, it suffers from a number of problems such as gradient vanishing, overfitting and curse of dimensionality. Some regularization techniques such as dropout are commonly used to tackle those problems.  There are also other types of architectures such as convolutional neural networks (CNNs), attention-based CNNs, Variational Autoencoder (VAE), Restricted Boltzmann Machines (RBMs) etc.~ that have much better performance for high-dimensional data. We review some more extended work on using more sophisticated neural network architectures for density estimation. \\

\hspace*{8mm} The early work using neural networks to perform density estimations was
extended by Iain Murray and his colleagues \cite{Larochelle:2011, Uria:2013, Uria:2014, Uria:2016}, Bengio and Bengio (2000) \cite{Bengio:2000}, and Gregor and LeCun (2011) \cite{GregorLeCun:2011}. Their approaches combine probabilistic graphical models (either directed or undirected) with neural networks such as restricted Boltzmann machines (RBMs)
and feed-forward neural networks. Among the seminal ideas in their work is the Neural Autoregressive Density Estimator (NADE) \cite{Larochelle:2011, Uria:2016}, which starts by factoring any $d$-dimensional distribution $p(\bx)$ $(\bx \in \Re^d)$ into conditional probabilities (for simplicity $\bx$ is assumed to be a binary vector): \\
\begin{equation}
p(\bx) = \prod_{k=1}^{d} p(x_k \mid \bx_{s<k})\,,
\end{equation}
where $\bx_{s<k}$ is the first $k-1$ subvector of the vector $\bx$. The autoregressive generative model of the data is defined by parameterizations of the $d$ conditional distributions $p(x_k \mid \bx_{s<k})$. Frey et al.~(1996) \cite{Frey:1996}
modeled the conditionals via log-linear logistic regressions, which yielded a competitive result.
Bengio and Bengio (2000) \cite{Bengio:2000}
extended the approach by modeling the conditionals via single-layer feed-forward neural networks. This gained some improvement in model performance at the cost of very large model complexity for high-dimensional datasets. In NADE \cite{Uria:2016}, 
they also model the conditionals using feed-forward neural networks via the following parameterizations: \\
\begin{align}
p(x_k =1 \mid \bx_{s<k}) &= \sigma (V_k \cdot \textbf{h}_k + b_k)\\
 \textbf{h}_k &= \sigma (W_{s<k} \cdot { \bx_{s<k}} + c)\,,
\end{align}
where $h$ is the hidden unit and $H$ is the number of hidden units. Then $W \in \Re^{H\times d}$ is the weight matrix for hidden units, $V^{d\times H}$, $b \in \Re^d$, and $c \in \Re^H$ are parameters associated with NADE models. Here
$\sigma (\theta) = \frac{1}{1 + e^{-\theta}}$ is the sigmoid activation function. The weight matrix $W$ and bias $c$ are shared among all the hidden layers $\textbf{h}_k$ having the same size
(shown in Figure \ref{fig:5}). This will reduce the total number of parameters from $O(Hd^2)$ to $O(Hd)$. Training the NADE can be done via maximum likelihood, or one can simply minimize the average negative log-likelihood:\\
\begin{equation}
-\frac{1}{N} \sum_{n=1}^N \log p(\bx^{(n)}) = -\frac{1}{N} \sum_{n=1}^{N}\sum_{k=1}^d \log p(x_k^{(n)} \mid \bx_{s<k}^{(n)})\,,
\end{equation} 
where $N$ is the number of training samples. Minimization of the objective function shown above can be readily achieved by stochastic (batch) gradient descent. Since there are $O(Hd)$ parameters in NADE, calculating $p(\bx)$ {costs} only $O(Hd)$, so the gradient of log-likelihood of training samples can also be calculated with the complexity $O(Hd)$. \\

\hspace*{8mm} The algorithms for calculating $p(\bx)$ and $-\nabla \log p(\bx)$ in Uria et al.~(2016) \cite{Uria:2016}
are illustrated in Algorithm 1.\\

\newpage
\begin{algorithm}[H]
	\caption{Computation of $p(\bx)$ and learning gradient in NADE (Uria et al. 2016) \cite{Uria:2016}} \label{alg3}
	\begin{tabbing}
		\enspace \textbf{Input:} training sample vector $\bx$ and ordering $s$ of the input dimensions \\
		\enspace \textbf{Output:} $p(\bx)$ and gradients $-\nabla \log p(\bx)$ on parameters \\
		\enspace \textbf{Computation} of $p(\bx)$ \\
		\enspace Set $\theta_1 \leftarrow c$\\
		\enspace Set $p(\bx) \leftarrow 1$ \\
		\enspace \textbf{For} $k$ from $1$ to $d$ \textbf{do} \\
		\qquad Set $\textbf{h}_k \leftarrow \sigma (\theta_k)$\\
		\qquad $p(x_k = 1 \mid \bx_{s<k}) \leftarrow \sigma (\textbf{V}_k \cdot \textbf{h}_k + b_k)$ \\
		\qquad $p(\bx) \leftarrow p(\bx) (p(x_k = 1 \mid \bx_{s<k})^{x_k} + (1 - p(x_k=1 \mid \bx_{s<k}))^{1 - x_k} )$ \\
		\qquad $\theta_{k+1} \leftarrow \theta_k + \textbf{W}_k \cdot x_k$ \\
		\enspace \textbf{end for}
		\\
		\\
		\enspace \textbf{Computation} of learning gradients of $-\log p(\bx)$ \\
		\enspace Set $\delta \theta_D \leftarrow 0$ \\
		\enspace Set $\delta c \leftarrow 0$ \\
		\enspace \textbf{for} $k$ from $d$ to $1$ \textbf{do} \\
		\qquad $\delta b_k \leftarrow (p(x_k = 1 \mid \bx_{s<k}) - x_k)$ \\
		\qquad $\delta \textbf{V}_{s<k} \leftarrow (p(x_k =1 \mid \bx_{s<k}) - x_k) \textbf{h}_k^\top$\\
		\qquad $\delta \textbf{h}_k \leftarrow (p(x_k = 1 \mid \bx_{s<k}) - x_k) \textbf{V}_k^\top$ \\
		\qquad $\delta c \leftarrow \delta c + \delta \textbf{h}_k \odot \textbf{h}_k \odot (1 - \textbf{h}_k)$ \\
		\qquad $\delta W_k \leftarrow \delta \theta_k x_k$ \\
		\qquad $\delta \theta_{k-1} \leftarrow \delta \theta_k + \delta \textbf{h}_k \odot \textbf{h}_k \odot (1 - \textbf{h}_k)$ \\
		\enspace \textbf{end for} \\
		\enspace \textbf{return} $p(\textbf{x})$, $\delta \textbf{b}$, $\delta \textbf{V}$, $\delta \textbf{c}$, $\delta \textbf{W}$
	\end{tabbing}
\end{algorithm}
\ \\

In their earlier work, Larochelle and Murray (2011 \cite{Larochelle:2011})
discussed the relationship between NADE and Restricted Boltzmann Machines (RBMs).  In RBMs, it is often intractable to compute the high-dimensional probability distribution because the partition function is intractable, even when the number of hidden units is
only moderately large. NADE approximates the partition function using mean field variational inference that makes the probability distribution completely tractable to compute. Uria et al.~(2013, 2016) \cite{Uria:2013, Uria:2016}
also extended NADE for real-valued variables. Interested readers should refer to the corresponding references for details. 
\begin{figure}[H]
	\centering
	\includegraphics[width=0.95\textwidth]{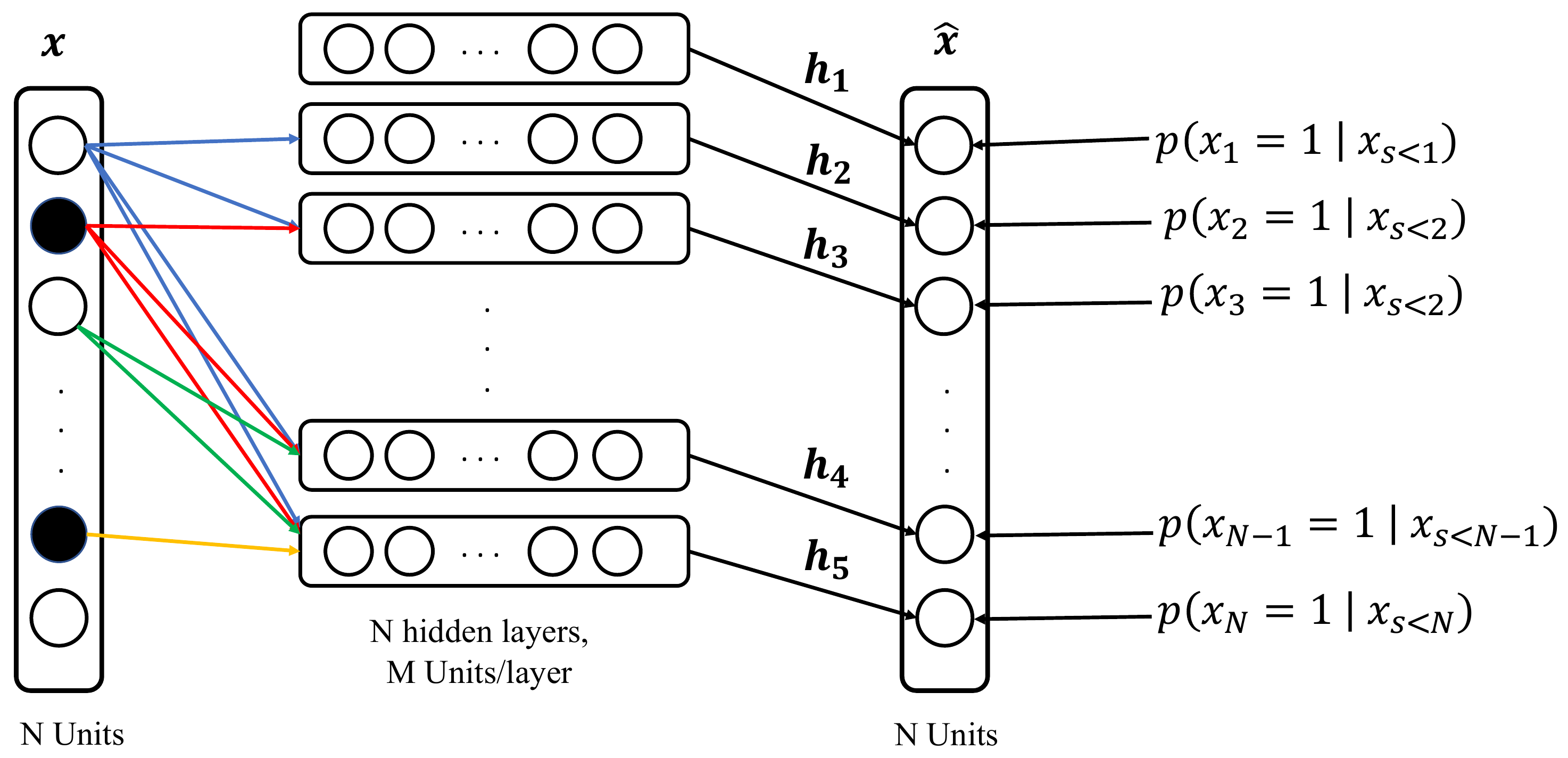}
	\caption{The architecture of a Neural Autoregressive Density Estimation (NADE) model. The input vector $\bx$ is a N-dimensional binary vector, units with value 0 are shown in color black, while the units with value 1 are shown in color white.  N input units represents the N dimensions in vector $\bx_o$. We basically model each conditional probability density $p (x_{d} = 1 \mid x_{s<d})$  using a single layer feed-forward neural network. There are N hidden layers to model N conditional probabilities. $\textbf{h}_\textbf{d}$ represents the d-th hidden layer ($d = 1, ..., N$). The output of each hidden layer is calculated via Equation 9. In this example, the vector $\hat{\bx}$ represents the output, and its dimensions $\hat{\bx}_i$ ($i = 1,.., N$) are the output of corresponding hidden layer $\textbf{h}_\textbf{i}$. Notice that each input unit connecting to the hidden layer through the weight-sharing scheme, which is highlighted in the figure with the same color.  \cite{Uria:2016} } %
	\label{fig:5}
\end{figure}

\hspace*{8mm} One of the limitations in NADE comes from its underlying assumption that the joint density function can be factorized into sequential conditional {densities}. In many real-world scenarios, the copula models of joint probability should be adopted. Interested readers can refer to Liu (2012) \cite{Liu:2012} and Dobra (2011) \cite{Dobra:2011}. \\

\hspace*{8mm}  Another critical drawback of NADE is that it is sensitive to the sequential order of the variables. For example, given $\bx \in \Re^d$, and let $\Pi _i (\bx), i = {1, ..., N}$ be a permutation order among elements $(x_1, ..., x_d)$ in $\bx$. A model with $\Pi_i (\bx)$ and $\Pi_j (\bx), j\neq i, j \in \{1, ..., N\}$ will likely to have different capability to learn certain density functions.  In practice, it is difficult to know which particular sequential order of the variables for the conditional factorization is optimal for the task. One solution to this problem is to train NADE with an sequential order of the variables at random, and combine the predictions from different sequential orders to form an ensemble model \cite{GermainICML2015, Uria:2014}. This requires $d$ sequential computations for estimating the density $p(\bx)$ because a hidden state needs to be updated sequentially for every variable. The computational disadvantage of the straightforward solution is not well-suited for large-scale parallel computation. Papamakarios and Murray (2017) \cite{Papamakarios: 2017} recently proposed a method called Masked Autoregressive Flow (MAF) that enables different sequential order of the variables in the conditional factorial and is well-suited for large parallel architecture such as GPUs. The proposed method can also perform density function estimation for real-valued variables. \\

\hspace*{8mm} Given an autoregressive model as follows: 

\begin{equation}
  p(\bx) = \prod_{k=1}^d p(x_k \mid \bx_{s < k})\,,
\end{equation}
each of the conditionals can be modeled as a single Gaussian distribution. To illustrate this, the $k^{th}$ conditional factor is given as follows: 

\begin{equation}
p(x_k \mid \bx_{s < k}) = \mathcal{N} \Big(x_k \mid \mu_k, (\exp(\beta_k))^2\Big)\,,
\end{equation}
where $\mu_k = f_{\mu_k} (\bx_{s < k})$ and $\beta_k = f_{\beta_k} (\bx_{s < k})$. 
      Note $f_{\mu_k}$ and $f_{\beta_k}$ are real-valued scalar functions that compute the mean and log standard deviation of the $i^{th}$ conditional distribution given all the ``previous" variables. The model also uses the vector of random variables $\textbf{u} = (u_1, ..., u_d)$ to generate data through the following recursive steps: 

\begin{equation}
x_k = u_k \exp(\beta_k) + \mu_k\,,
\end{equation}
where $\mu_k = f_{\mu_k} (\bx_{s<k}),  \beta_k = f_{\beta_k} (\bx_{s<k})$, and $u_k \sim \mathcal{N} (0, 1)$. \\
\hspace*{8mm} The MAF model stems from Normalizing flows \cite{Rezenda: 2015}, which expresses the joint density function $p(\bx)$ through the invertible, differentiable function $f$ of a low-level density $q_u (\textbf{u})$. It is straightforward to see that $\bx = f(\textbf{u})$ where $\textbf{u} \sim q_u (\textbf{u})$. The density $q_u (\textbf{u})$ should be carefully chosen so that it is easy to be evaluated at any variable value of $\textbf{u}$ (e.g. standard Gaussian). Under the theorem of invertible functions, the joint density $p(\bx)$ can be expressed as: 
\begin{equation}
p(\bx) = q_u\big(f^{-1} (\bx)\big) \cdot \Bigl\vert \det(\frac{\partial f^{-1}}{\partial \bx}) \Bigr\vert \,.
\end{equation}

In order to compute the density $p(\bx)$, the function $f$ has to be easily invertible, and the determinant of the Jacobian should be easy to compute. Go back to the MAF, $\bx = f(\textbf{u})$, where $\textbf{u} \sim \mathcal{N} (0, \textbf{I})$. Then given a data point $\bx \in \Re^d$, the random number $\textbf{u}$ will be derived from the following steps: 
\begin{equation}
u_k = (x_k - \mu_k) \exp(- \beta_k), \ \  \mu_k = f_{\mu_k} (\bx_{s<k}), \ \ \beta_k = f_{\beta_k} (\bx_{s<k})\,.
\end{equation} 
In the Autoregressive model, the Jacobian of $f^{-1}$ has a triangular structure, so the determinant is: 
\begin{equation}
\Bigl\vert \det(\frac{\partial f^{-1}}{\partial \bx}) \Bigr\vert = \exp \left(\sum_k \beta_k\right)\,,
\end{equation}
where $\beta_k = f_{\beta_k} (\bx_{s<k})$. 

\hspace*{8mm} The density $p(\bx)$ can be obtained by substituting Equations (15) and (16) into Equation (14), so it can also be interpreted as a normalizing flow \cite{Kingman:2016}.  The implementation of the set of functions $\{f_{\mu_k}, f_{\beta_k}\}$ with masking borrows the idea from the Masked Autoencoder Density Estimation (MADE) \cite{GermainICML2015}. MADE is simply a feed-forward neural network which takes the input data $\bx$ and outputs mean $\mu_k$ and variance parameter $\beta_k$ for all $k$ with a single round of pass.  In MAF,  the weight matrices of MADE are multiplied by the binary masks to ensure that the autoregressive properties are well-maintained.  In other words, MAF uses the MADE with Gaussian conditionals as the building layer of the flow. The flow in MAF is interpreted as a flow of autoregressive models through stacking multiple autoregressive model instances, which improves the model fit. The reason to use the masking approach is that it enables transforming the input data $\bx$ into the random number $\textbf{u}$ and calculating the density $p(\bx)$ by finishing one round pass through the flow, instead of doing recursive calculations as in Equation (15). \\

\hspace*{8mm} MAF can potentially learn complex density functions for high-dimensional data. Using images from MNIST datasets as examples ({note that MNIST datasets stand for Modified National Institute of Standards and Technology datasets, which is a large database of handwritten digits widely used in training many machine learning algorithms for image processing}), MAF can successfully learn the complex density functions the MNIST images, and generate images which capture the underlying patterns of the real images;  see Figure 2.  However, compared to modern image-based generative models such as PixelCNN++ \cite{Kingma:2017}, RealNVP \cite{SBengio:2017} or CycleGAN \cite{JunYanZhu:2017}, the MAF-generated images lack the fidelity provided by those models. But MAF was originally designed as a general-purpose density estimator rather than a domain-specific generative model. Interested readers can refer to those original papers for details.  

\begin{figure}[H]
	\centering
	\includegraphics[width=0.95\textwidth]{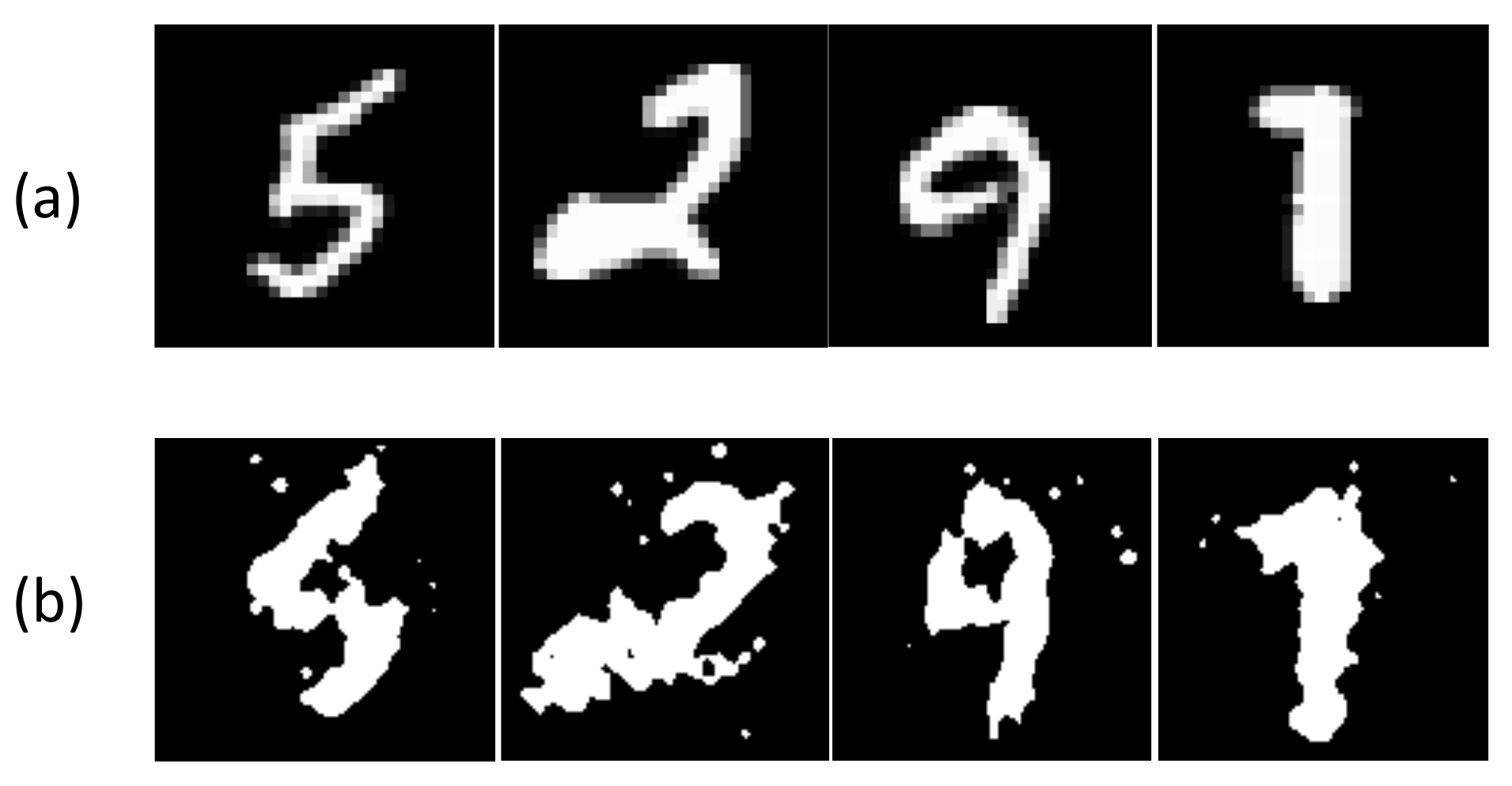}
	\caption{Real Images and Generated Images by MAF from MNIST datasets. (a) Real Images from MNIST datasets; (b) Generated Images by MAF through MNIST datasets. \cite{Papamakarios: 2017} }
	\label{fig:6}
\end{figure}

\subsection*{Density Estimation Based on Adaptive Data partitioning}
It is widely accepted that kernel-based nonparametric density estimation is computationally intense, and it also suffers from the problem of ``curse of dimensionality." Essentially as the dimension increases, the number of data points needed to get a reliable density estimator grows exponentially. How can we come up with a reliable density estimator for high-dimensional data with limited amounts of data and computational resources? \\

\hspace*{8mm} Wong and his colleagues have come up with some novel methods to address this problem (Wong and Ma, 2010; Lu and Wong, 2013; Li and Wong, 2016) \cite{MaWong:2010, LuoWong:2013, LiWong:2016}. The basic idea is to treat the multivariate density estimation as a nonparametric Bayesian problem. Suppose $X$ is a random variable on a space $\Omega$ ($X \in \Re^d$), their distribution $Q$ is unknown but it is assumed to be drawn from a prior distribution $\pi$. The posterior $P(Q \mid X)$ can be calculated by $\pi (Q) Q(X)$. Choosing prior distributions should follow Ferguson's criteria that 1) there should be a large support for the prior and 2) the posterior distribution should be tractable. Although the commonly-adopted Dirichlet process satisfies the Ferguson's criteria, the corresponding posterior does not possess a density function. So instead they considered a class of piecewise constant density functions over partitions of the data space: $Q(X) = \sum_{i=1}^m c_i I_{A_i}(X)$, where
$m$ is the number of partitions, $A_i$ is the $i$th partition, and $I_{A_i}$ is the indicator function. As usual,
$I_{A_i}(X) = 1$ indicates that $X$ falls into the $i$th partition; otherwise
$I_{A_i}(X) = 0$.  From the piecewise constant prior distribution over partitions, they derived a closed-form marginal posterior distributions for corresponding partitions. The inference on the partitions is achieved by their proposed algorithm that they called ``Bayesian sequential partitioning (BSP)." The basic idea is that in each iteration, binary partitioning on one of the subregions and dimensions of the data domain $\Re^d$ is performed. Then the posterior distribution of the corresponding partitions,
$\pi^0(\cdot)$, is calculated and used to assign ``scores" to each partition. Since the closed-form posterior distribution over partitions is available, the inference of partitions can be done via Markov Chain Monte Carlo (MCMC) or Sequential Importance Sampling (SIS). \\

\hspace*{8mm} Figure \ref{fig:6} illustrates the path for BSP while Figure \ref{fig:7} shows how the Sequential Importance Sampling works. In the space of partitioning paths, the density is defined as $q(g_t)$, which is proportional to the posterior probability for the partition generated from the partitioning path $g_t = (d_1, d_2, ..., d_t)$, where each $d_j, j \in \{1, 2, ..., t\}$ represents the partition decision at level $j$ for dividing a subregion in the partition generated by $g_{t-1} = (d_1, d_2, ..., d_{t-1})$.  Let $x_t$ denote the partition generated from $g_t$. Since there are potentially many partition paths that lead to the same partition $x_t$, the author introduced the notation
$\Lambda (x_t)$ to present the number of unique paths that lead to the same partition $x_t$. If the partition paths $g_t^{(1)}, g_t^{(2)}, ..., g_t^{(m)}$ are generated from the probability distribution of paths $q(\cdot)$, the corresponding partitions are $x_t^{(1)}, x_t^{(2)}, ..., x_t^{(m)}$.  Here, each $x_t^{(i)}$ is the
partition generated from the path
$g_t^{(i)}$,
where $x_t^{(i)} = x(g_t^{(i)})$, can be treated as a weighted sample with the posterior distribution $\pi^0 (g_t)$ for the partition and the weights $w_i = 1 / \Lambda(x_t^{(i)}), i \in \{1,2, ..., m\}$.  Since $q(g_t) \propto \pi^0 (x(g_t))$ and the partition path is sequentially constructed, the weighted samples of partition paths can be generated by using Sequential Importance Sampling \cite{Gordon:1993, KongLiuWong:1994, Liu:2001} as shown in Figure \ref{fig:7}. \\

\hspace*{8mm} They calculated the partition score, which is just the logarithm of the posterior probability (as a function of the number of partitions $t$) and the KL divergence between the estimated density and the true density as a function of $t$ for a simulated mixture Gaussian Distribution. Their results indicate that the partition score tracks the KL divergence with higher partition scores corresponding to smaller KL divergences;
see Wong \cite{Wong:2014}. \\
\begin{figure}[H]
	\centering
	\includegraphics[width=0.95\textwidth]{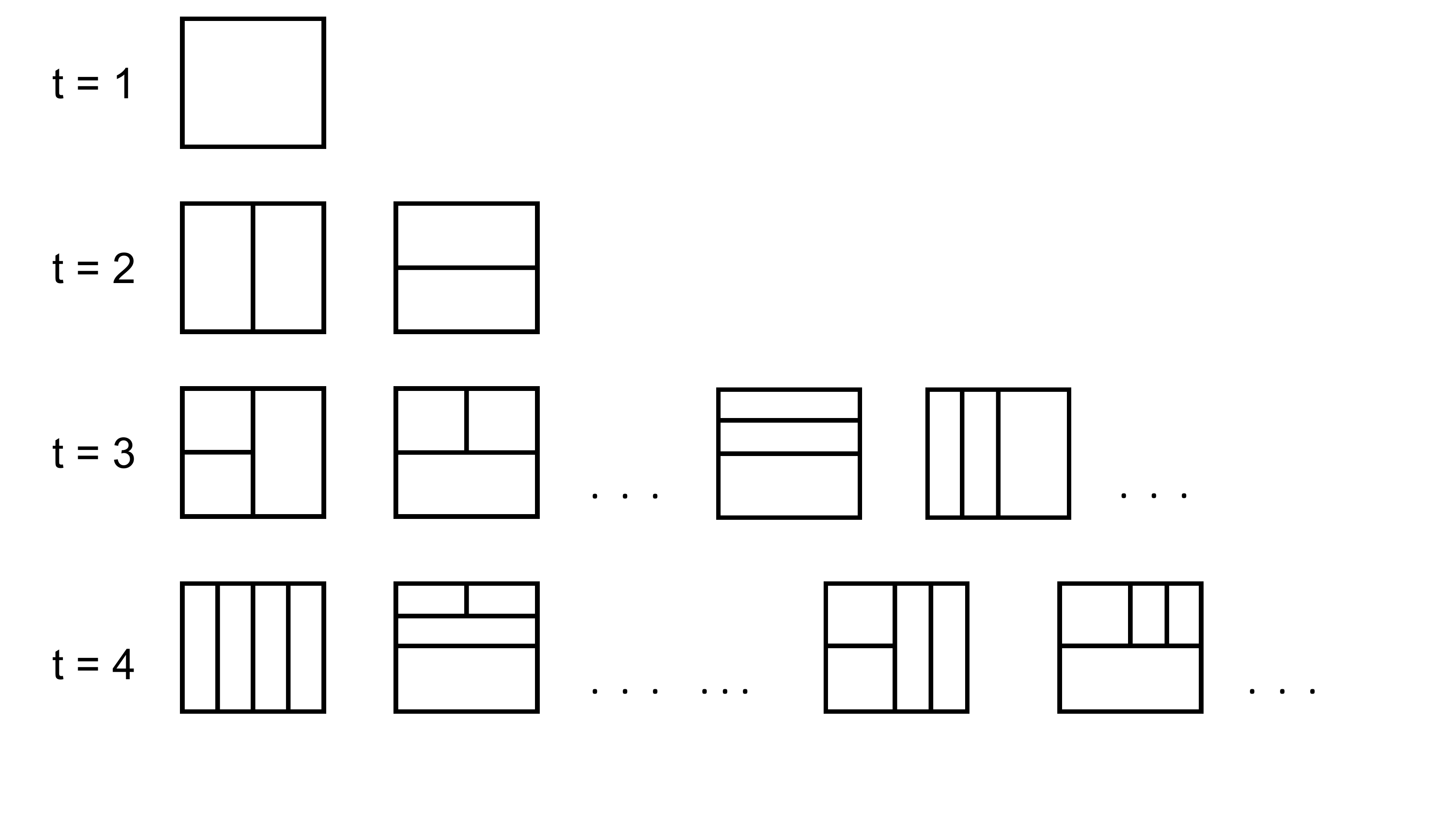}
	\caption{Recursive Sequential Binary Partitioning, where $t =1, 2, 3, 4$ represent the level of partition, and the partition is performed sequentially. At each level, there are a variety of different ways to perform binary partition; (from Wong (2014) \cite{Wong:2014}).}
	\label{fig:6}
\end{figure}

\begin{figure}[H]
	\centering
	\includegraphics[width=0.95\textwidth]{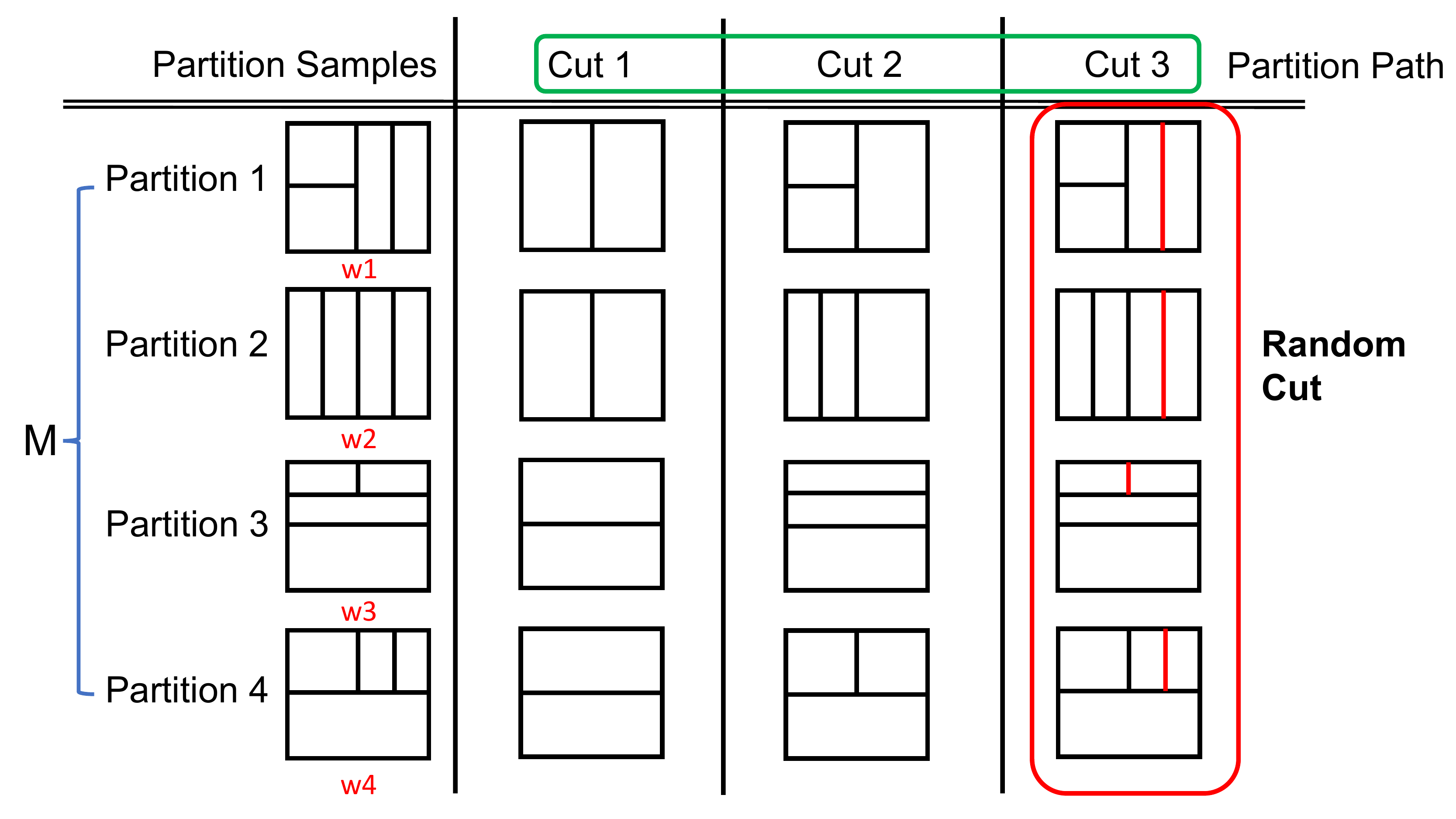}
	\caption{Sequential Importance Sampling (SIS) to generate weighted samples of partition paths. Here 4 partition samples are illustrated, and their corresponding weights are $w_1, w_2, w_3, w_4$, respectively; (from Wong (2014) \cite{Wong:2014}).}
	\label{fig:7}
\end{figure}


\hspace*{8mm} {Li et al. (2016) \cite{LiWong:2016}} further extended the BSP method, leading to a more flexible and intuitive algorithm for multivariate density estimation. They use
a greedy algorithm to determine the splitting point by picking the maximum gap,
$g_{jk}$,
where $j$ represents $j$th dimension and $k$ represents the $k$th splitting point along any of the dimensions (any dimension will be divided into $m$ equally-spaced bins). Given $n$ data points $X_n = {x_1, x_2, ..., x_n}$ transformed into $[0, 1]^d$,
$g_{jk} = \, \big|\, (\frac{1}{n}) \sum_{i=1}^n \textbf{1} (x_{ij} < a_j + (b_j - a_j)k/m) - k/m \,\big|\,,$ for $k=1,2,...,m-1$. There are $(m-1)d$ recorded in total. At each iteration the {maximum gap} will be picked and the corresponding splitting point will be selected. The algorithm keeps iterating until the maximum gap falls below a predetermined
threshold. The algorithm is computationally efficient and has been proven to converge. For more technical details readers are encouraged to look into the original reference. \\

\hspace*{8mm} There are several potential limitations of BSP algorithm and its extended version: 1) data partitioning has to be sequentially performed, so it might take a long time to converge when number of dimensions gets high; 2) the resulting piecewise constant density function is discontinuous, in which case the edge problem from multivariate histogram estimators will arise, so the density estimation might be rather biased; and 3) for the BSP, it is not always appropriate to choose a prior distribution that is piecewise constant, so that different priors might be needed, in which case the posterior distribution might not possess a closed-form expression. One potential way to improve the statistical bias is by using a multivariate frequency polygon (Hjort, \cite{Hjort:1986}). Another potential way is to identify clusters via spectral clustering \cite{ShiMalik:2000, MarinaShi:2000, Dhillon:2004}, and then perform discrete convolution over the entire data domain, which is an ongoing project by the current authors and will not be covered by this article. 

\subsection*{Projection Pursuit Density Estimation (PPDE)}

One of the first ``advanced'' algorithms for finding structure (read: clusters)
in data was devised by Friedman and Tukey \cite{Friedman:1974}.  A nonparametric
criterion was presented that used to find low-dimensional projections with
``interesting'' views.  They estimated that interactive examination of all projections
would quickly become inefficient for a human, and so they devised ``cognostics"
or criteria that would allow
the computer to perform an extensive search without human intervention.
Interesting projections (local optima in the cognostics space) could then be presented to
the human user.  They called their algorithm {\it projection pursuit} (see also
\cite{Friedman:1987}). 

\hspace*{8mm} A number of extensions have been investigated subsequently, namely, projection
pursuit regression \cite{Friedman:1981} and projection pursuit
density estimation (PPDE) \cite{Friedman:1984}. The motivation behind the PPDE algorithm is to tackle the 
poor performance of multivariate kernel density estimation when it comes to high-dimensional data, because extremely large sample sizes are needed to achieve the level of numerical accuracy available in low dimensions. The PPDE algorithm uses an iterative procedure to find the interesting subspace, which is spanned by a few significant components. The detailed procedure is outlined in Algorithm 2. 

\newpage
\begin{algorithm}[H]
	\caption{Projection Pursuit Density Estimation (PPDE) Algorithm} \label{alg3}
	\begin{tabbing}
		\enspace \textbf{Input:} Observed Data, $L = \{x_i, i=1,2,...,n\}, x_i \in\Re^d$, Scale the data \\
		\enspace  to have mean $\textbf{0}$ and covariance matrix $\textbf{I}_d$ \\
		\enspace \textbf{Initialization}: Choose $\hat{p}^{(0)}$ to be an initial density estimate of $p (\bx)$, \\ \hspace{1.0 in} usually it is taken to be standard multivariate Gaussian\\
		\enspace \textbf{For} $k=1,2, ...$ \\
		\qquad 1. { Find the direction $\textbf{c}_k \in \Re^d$ for which the marginal (model) $p_{\textbf{c}_k} (\bx)$ along}\\
		\qquad \qquad { $\textbf{c}_k$ differs most from the current estimated data marginal $\hat{p}^{(k-1)}_{\textbf{c}_k}$ along $\textbf{c}_k$} \\
		\qquad \qquad {the choice of $\textbf{c}_k$ generally will not be unique)}\\
		\qquad 2. Given $\textbf{c}_k$, calculate the univariate ``ridge" function: \\
		\qquad \qquad $t_k (\textbf{c}_k^{\tau} \bx) = \frac{p_{\textbf{c}_k} (\textbf{c}_k^{\tau} \bx)}{\hat{p}_{\textbf{c}_k} (\textbf{c}_k^{\tau} \bx)}$ \\
		\qquad 3. Update the previous estimate of the probability density function:  \\
		\qquad \qquad $\hat{p}^{(k)}(\bx) = \hat{p}^{(k-1)} (\bx) t_k (\textbf{c}_k^{\tau} \bx)$ \\
		\enspace \textbf{End For}
	\end{tabbing}
\end{algorithm}
\vspace{.3in}

Just to clarify the notation used in Algorithm 2,  the vectors $\{ \textbf{c}_k \}$ are unit-length directions in $\Re^r$, and the ridge functions $\{t_k \}$ are constructed so that $\hat{p}^{(k)}$  converges to $p$ numerically as $k \rightarrow \infty$. The number of iterations $k$ serves as a smoothing parameter, and the iteration ceases when the stopping rule determines that the estimation bias is balanced out against the estimation variance. Computation of the ridge function $t_k(\textbf{c}_k^{\tau} \bx)$ can be done via two steps: 1) given $\textbf{c}_k$, project the sample data along the direction $\textbf{c}_k$, thus obtaining $q_i = \textbf{c}_k^{\tau} x_i, i=1,2,...,n$; and 2) compute a kernel density estimate from the projected data $\{q_i \}$. Computing the $\hat{p}_{\textbf{c}_k}$ is done via Monte Carlo sampling followed by a kernel density estimation. Alternative smoothing methods include cubic spline functions \cite{Friedman:1984} and the average shifted histograms \cite{Jee:1987}. \\

\hspace*{8mm} The specific use of projection pursuit for finding clusters has been
investigated recently by Tasoulis, et al. \cite{Tasoulis:2012}.
However, the underlying technology has
not seen the rapid development seen by algorithms and has been limited
to projection to a few dimensions.  Given that projection tends to reduce
the number of modes (clusters) by combination (overlap), we do not pursue this further.
A nice review is provided by Jones and Sibson \cite{Jones:1987}.


\section*{APPLICATIONS of NONPARAMETRIC DENSITY ESTIMATION: MODAL CLUSTERING}

{In this section, we would like to discuss one of the important application domains of nonparametric density estimation, which is modal clustering. Modal clustering is a clustering approach that determines the clusters of data through identifying modes in the probability density function. Each mode is then associated with a cluster. The technical challenge is to discover all the true modes in the density function through data-driven approach. }
Good and Gaskin (1980) \cite{Good:1980}
pioneered the way of using nonparametric density estimations as a powerful tool for discovering {modes} and bumps. They used a Fourier Series representation with thousands of terms to fit the probability density function in $\Re^1$ and identify the {modes}, which is rather impressive given the computing resources at the time. { Here we want to show the general mathematical insight of mode finding and modal clustering. In order to facilitate the discussion, we use the kernel density estimator as an exemplary density estimation algorithm.} The basic spherical
kernel estimator may be written compactly as:
\begin{equation}
\hat{f}(\bx) = \frac{1}{nh^d} \sum_{i=1}^{n} K(\frac{\bx - \bx_i}{h}) = \frac{1}{n} \sum_{i=1}^{n} K_h (\bx - \bx_i)\,,
\end{equation}
where $\bx_i \in \Re^d\ (i = 1, 2, ...,n) $ are data points and $h$ is the smoothing parameter, which is applied to each dimension.
We usually choose the kernel $K$ to be a standard Gaussian density, $\mbox{MN}(\bzero,\bI_d)$
While a more general covariance matrix may be selected,
it is equivalent to linearly transforming the data in a certain manner, so
that this kernel is fully general.

\hspace*{8mm} An important result discovered by Silverman (1981) \cite{Silverman:1981}
was that with the Gaussian kernel, the number of modes decreases monotonically as the smoothing parameter $h$ increases,
at least in one dimension;  however, the result does not extend to higher dimensions;
	see Figure \ref{fig:3}.
The necessity (Silverman's paper showed sufficiency)
to use Gaussian kernel for mode clustering in 1-D was proven by
Babaud et al.~(1994) \cite{Babaud:1994}.
Minnotte and Scott (1993) \cite{Minnotte:1993}
created a tree-like graphical summary that captured all of the modes of a Gaussian kernel estimate as a function of the bandwidth $h$. It is called the
``mode tree" in their original paper.
Figure \ref{fig:2} displays the mode tree on the Old Faithful Geyser dataset and may be compared with the dendrogram hierarchical clustering tree on the same dataset. Since there is no probabilistic model for hierarchical clustering, it is difficult to analyze or make any probabilistic statement about features in the dendrogram. By contrast the mode tree is a powerful visualization tool for modes.
Minnotte was able to test the veracity of individual modes by using
the structure of the mode tree;  see  Minnotte and Scott (1993) \cite{Minnotte:1993} and
Minnotte (1997) \cite{Minnotte:1997}.
Since no single choice of bandwidth $h$ is likely to show all the potential modes at the same time, assessing the modes or the number of clusters is extremely challenging. In that sense the bootstrap is an appealing means to tackle this problem, due to its generality and lack of specific assumptions. Likewise since there is no probability model for the dendrogram clustering tree, it is difficult to assess dendrogram tree representation across bootstrap samples, since it is not obvious how to compare pairs of clustering trees due to the change in labels. The mode tree
solves this problem. It is also worth noting that
Silverman (1981) \cite{Silverman:1981}
suggests a conservative test of the null hypothesis in the univariate case that the unknown density has at most $k$ modes with a certain bandwidths $h$, which he called ``critical bandwidths."
These are defined by an additional mode about
to appear when the bandwidth is decreased further. This is where the horizontal dashed lines appear in the mode tree in Figure \ref{fig:2}. 
\begin{figure}[H]
	\centering
	\includegraphics[width=0.95\textwidth]{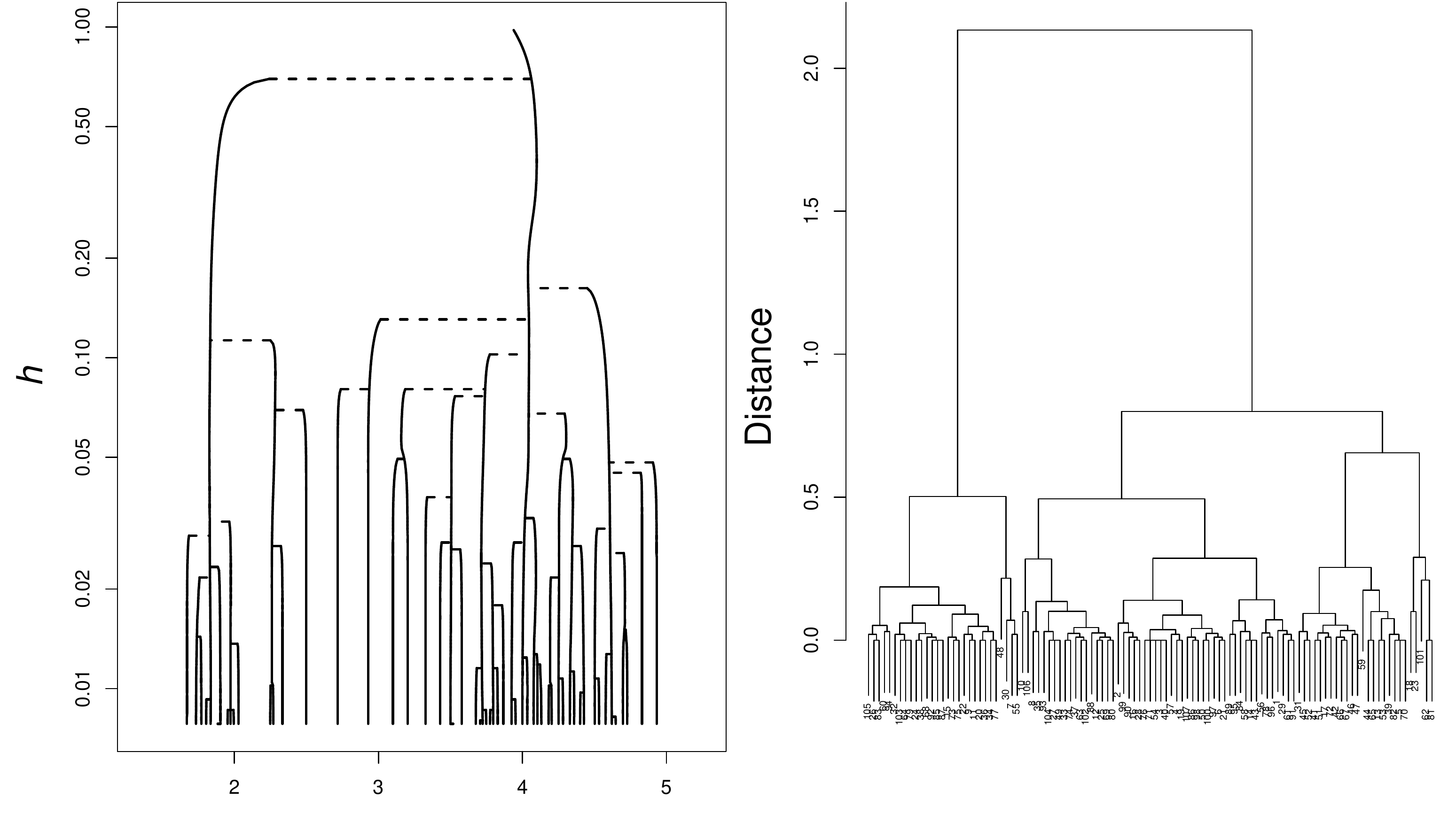}
	\caption{Illustration of the mode tree (left) and the dendrogram clustering tree (right) of the geyser eruption times dataset. Notice that the dendrogram (right) is created by hierarchical clustering based on the average linkage between modes. }
	\label{fig:2}
\end{figure}
\hspace*{8mm} In Silverman's study, he suggested counting the number of modes at those critical bandwidths across many bootstrap samples, assessing the veracity of that mode count by the distribution of the mode count in the resamples. It was noted by
Matthews (1983) \cite{Matthews:1983} that this approach might not work for complex densities. To be specific, if there is a true mode at a relatively low height that requires a small bandwidth to properly resolve, other taller modes may split into many noisy modes at the appropriate critical bandwidth, which make it very likely to mask the smaller mode across bootstrap samples (Scott, 2015 \cite{Scott:2015}).
Also the final counting of modes will be influenced by any outliers because in the bootstrap sample the outliers show up with the probability of $1- e^{-1} \approx 0.632$ of the time.
To rephrase, a single ``good" choice of the bandwidth $h$ is likely to both undersmooth and oversmooth in certain regions of the density, so that the count of modes is not very precise. \\ 

\hspace*{8mm} {Minnotte (1997) \cite{Minnotte:1997}}
successfully showed that by testing the individual modes at critical bandwidths along the branch of the mode tree, one can appropriately evaluate the modes locally. The idea is quite intuitive since nonparametric density estimation
enjoys its success largely by
being a local estimator.  Other tree-like assessments including SiZer (\cite{Marron:1999}, \cite{Marron:2000})
which is based on a scale-space representation.
Erasto and Holmstrom (2005) \cite{Erasto:2005}
proposed a Bayesian version of SiZer, and {Minnotte, et al.~(1998)} \cite{Minnotte:1998}
proposed a ``mode forest,'' which is an enhanced version of mode-tree to present a collection of bootstrapped mode trees. Another approach which was proposed for mode assessments is to look at contours around a mode and compute the ``excess mass" in that region (Muller and Sawitzki (1991) \cite{Muller:1991}
and Mammen et al.~(1994) \cite{Mammen:1994}).
In high-dimensional problems where $d >1$, the Hessian might be indefinite in some regions with a mixture of positive and negative eigenvalues. In that case the analysis becomes quite complicated
and special cares have to be taken; see Marchette, et al.~\cite{MarchetteWegman:1996} and Wong (1993) \cite{Wong:1993}. \\

\hspace*{8mm} In one dimension, there is a sufficient and necessary condition supporting that the number of modes is monotone for Gaussian kernel. The natural question is how the number of modes change in high-dimensions with the bandwidth $h$.
Scott and Szewczyk (1997) \cite{Scott:1997}
provided a counterexample that is shown in Figure \ref{fig:3};
see Scott (2015) \cite{Scott:2015}.
\begin{figure}[H]
	\centering
	\includegraphics[width=0.55\textwidth]{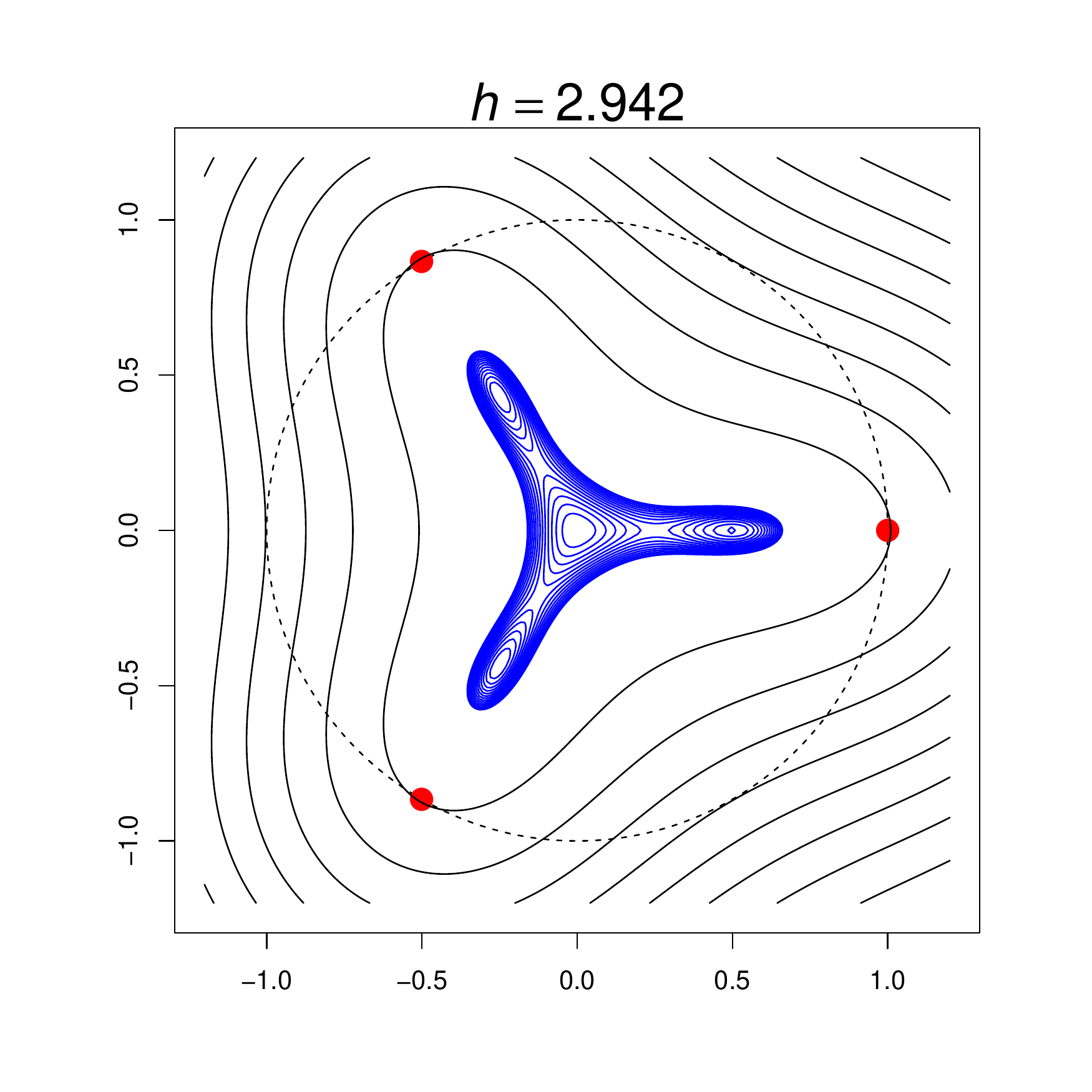}
	\caption{{\color{black} Example of 4 modes but only 3 data points (shown
			in red at vertices of an equilateral triangle) in two dimensions.
			The surface is very flat, which is highlighted by the blue contours around the origin.
			See the text for further details.}}
	\label{fig:3}
\end{figure}
\hspace*{8mm} Figure \ref{fig:3} shows three data points at the vertices of an equilateral triangle, whose kernel density estimator generates three modes if $0 < h < 2.8248$ and one mode if $h > 2.9565$.
However, in the very narrow range of bandwidths $2.8248 < h < 2.9565$, a fourth mode
appears at the origin, one more than the true number of components.
Figure \ref{fig:3} shows where the four modes are located and all are clearly visible.
In their paper a circular covariance matrix is used,
which corresponds to a mixture of three equally weighted Gaussian densities. Their result indicates that in high-dimensions, monotonicity also does not hold and that the range of bandwidths where this holds grows slightly with dimension.
Other authors have found other counterexamples for points on a right angle rather than
the regular mesh pattern here and unequal covariance matrices \cite{Carreira:2003}. \\

\hspace*{8mm} Extensive discussions of the multivariate versions of the mode tree
have been
done by Minnotte and Scott (1993) \cite{Minnotte:1993}
and Klemela (2008, 2009) \cite{Klemela:2008, Klemela:2009}.
Figure \ref{fig:4} illustrates the bivariate mode tree for the lagged geyser dataset.
The trimodal feature is clearly visible.
In more than 2 dimensions, placing $n=d+1$ data points at the vertices of a regular polytope (the regular tetrahedron in $\Re^3$ for example) and using the $MN(\bzero,\bI_d)$ kernel, we observe either 1, $d+1$, or $d+2$ modes.  The range of bandwidths $h$ where the ``phantom mode'' at the origin is observed increases as $d$ grows by empirical observation.  In our opinion, the possibility of phantom modes has little impact on clustering, but must be accounted for when programming and evaluating the mode tree.  Assuming monotonicity can defeat a code when phantom modes appear.

\begin{figure}[H]
	\centering
	\includegraphics[width=0.45\textwidth]{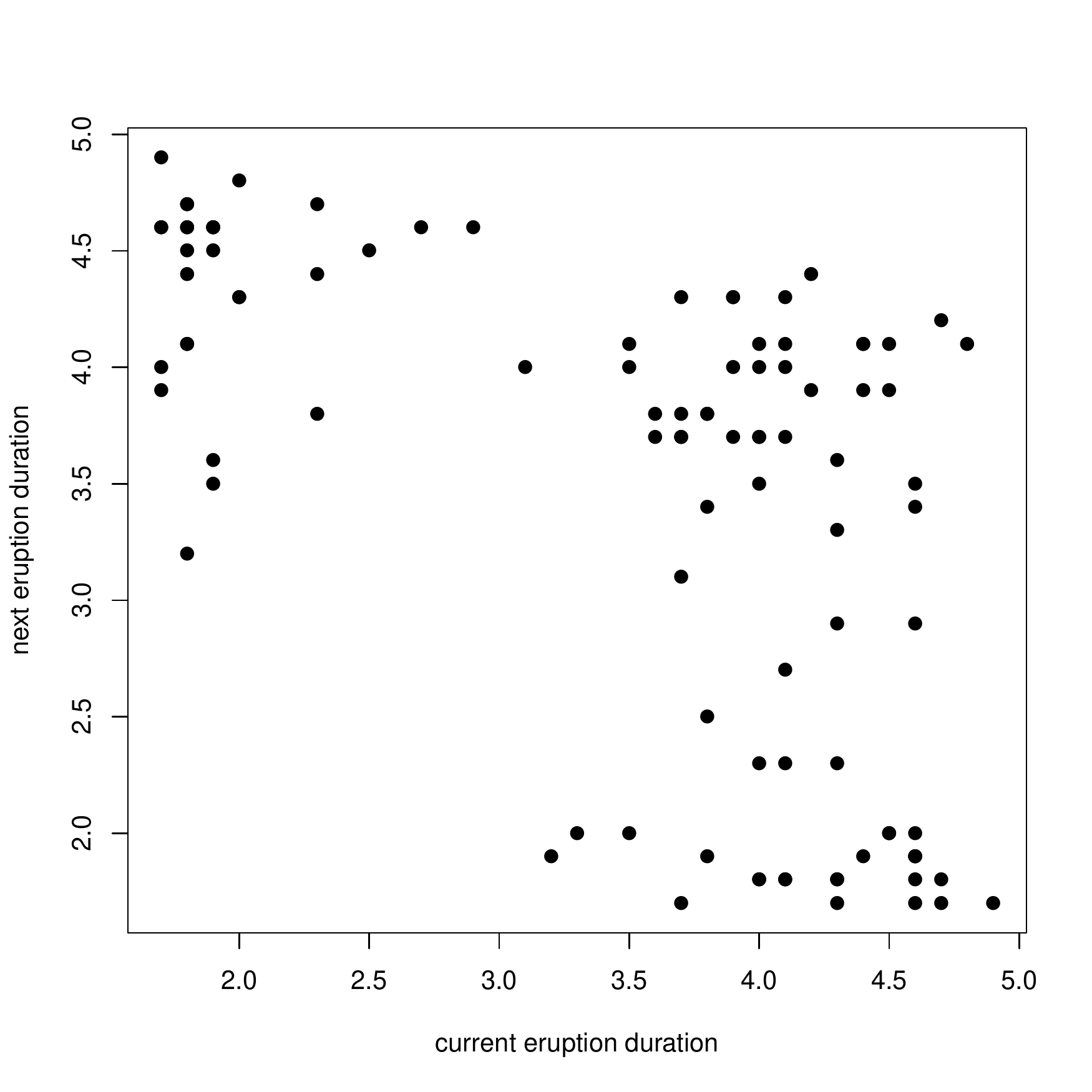}
	\includegraphics[width=0.45\textwidth]{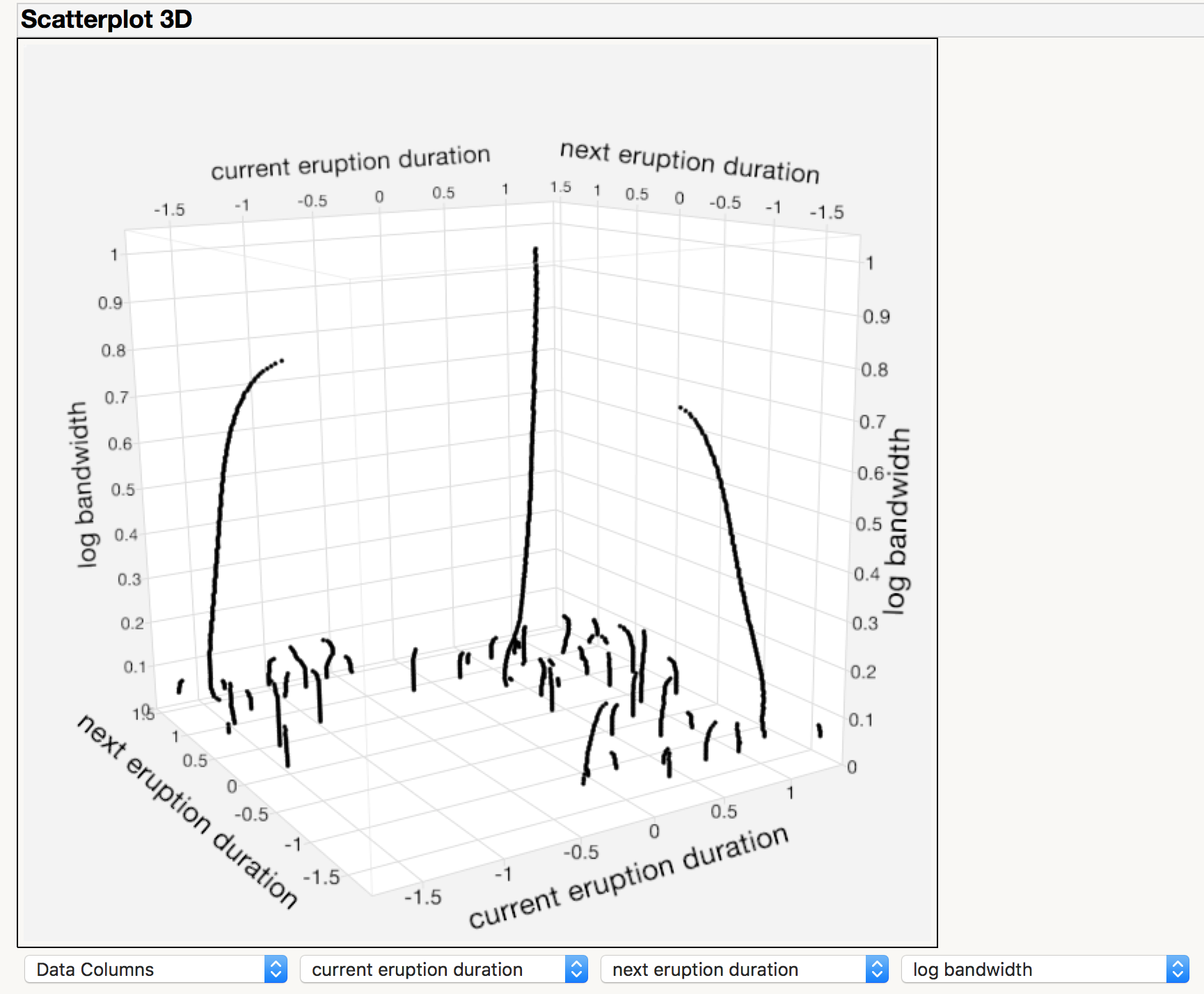}
	\caption{{\color{black} Bivariate scatterdiagram and bivariate
			mode tree for lagged geyser dataset.  Kernel
			estimates for 201 choices of the logarithm
			of bandwidth $h$ (scaled to $(0,1)$ were computed
			and the sample modes located.  The data have three obvious clusters,
			which are visible in the scatterdiagram as well as the three
			long modal traces in the right frame.}}
	\label{fig:4}
\end{figure}

\hspace*{8mm} There are several challenges associated with mode clustering, especially in high dimensions {($d\ge4$)}. First of all, by the hard assignment of data points, it is difficult to evaluate the uncertainty of how well the data points are being clustered. How to visualize clusters in high dimensions ($d\ge4$) also remains a difficult problem;
see the ideas in Klemel\"a \cite{Klemela:2008, Klemela:2009}
as well as Stuetzle and Nugent
\cite{Stuetzle:2003, Stuetzle:2010}.
As discussed before, the number of modes is heavily dictated by bandwidth $h$, and identifying the appropriate bandwidth $h$ for the kernel density estimator is not trivial.
Using only one bandwidth everywhere with a kernel
estimate is usually far from adequate as regions of oversmoothing
and undersmoothing are inevitable.  Thus, in high dimensions one cannot (entirely)
avoid the likelihood that noisy modes will appear, even if local smoothing
is attempted.  How to assess those small noisy modes as well as
missing or extra modes, and how to further denoise those modes are complicated questions, requiring sophisticated statistical analysis and further research.  However,
Chen et al.~(2016) \cite{Chen:2016} has
proposed a solution to all of those problems that leads to a complete and reliable approach for modal clustering. Notably they provided a soft assignment method that is able to capture the uncertainty of clustering, and define a measure of connectivity among clusters.
They also proposed an estimate for that measure. Their method was proved to be consistent and is able to provide an enhanced capability for mode clustering in high-dimensions. Wasserman (2018) \cite{Wasserman:2018}
discuss mode clustering in high dimensions in the context of topological data analysis, which represents a generalized collection of statistical methods that identify intrinsic structures of data by use of ideas of topology. Interested readers can also refer to Chen et al.~(2016) \cite{Chen:2016}.  We note, however, that they usually use
a single bandwidth globally, so the results will be asymptotic to the globalized distributions unless the
modes or features have nearly the same height and shape. \\

\hspace*{8mm} We conclude by observing that Ray and Lindsay \cite{Ray:2005} have
given an elegant algorithm for finding all modes as $h$ varies by following
the so-called ``density ridges."  These can also serve as a visualization tool,
which are rather {different than} those of Klemela \cite{Klemela:2008}.  For
$d>2$, the Minnotte-Scott mode tree reverts to the
dendrogram-like appearance as
in {Figure 5}.  The coalescence of adjacent (neighboring) modes may be determined by using the Ray-Lindsay algorithm. {Interested readers can also refer to the work of Minnotte(2010) which uses high-order variants on kernel density estimation to test multimodality \cite{Minnotte:2010}}. 
Finally, a partial survey of software available for modal clustering may
be found in the Menardi survey \cite{Menardi:2016}.
\\



\section*{SUMMARY}
{Nonparametric density estimation is an active research field in machine learning and statistics. Conventional method such as Kernel Density Estimation (KDE) performs poorly for high-dimensional data ($d > 3$). For the real-world problems, ideally we want to have reliable density estimators for $3 < d \leq 50$. In this paper we reviewed some selected nonparametric density estimation algorithms which could potentially tackle high-dimensional problems.} {On the application side,} modal clustering based on nonparametric density estimations enjoys high flexibility, adaptivity and performs well in a wide range of scenarios. {The reviewed multivariate density estimation algorithms provide powerful building blocks for modal clustering.}  They are also able to compensate for some of the limitations of KDE.  Future research should focus on developing more efficient, scalable, and reliable density estimation algorithms that
work effectively in high dimensions.  These  algorithms ideally should lead to density functions that are as smooth as possible. They should also exhibit the property of effective local smoothing to minimize the likelihood of false or missing modes as correctly as possible.
\end{justify}


\section*{ACKNOWLEDGMENTS}

The first author acknowledges the financial support from NSF and Rice University. 
The authors would also like to thank the two referees for helpful suggestions that refocused the article.

\end{document}